\documentclass[conference]{IEEEtran}
\IEEEoverridecommandlockouts
\usepackage[table]{xcolor}
\usepackage{cite}
\usepackage{amsmath,amssymb,amsfonts}
\usepackage{algorithmic}
\usepackage{graphicx}
\usepackage{lipsum}
\usepackage{capt-of}
\usepackage{cuted} 
\usepackage{comment}
\usepackage{svg}
\usepackage{textcomp}
\usepackage{booktabs}
\usepackage{ifthen}
\usepackage{multirow}
\usepackage{booktabs}
\newboolean{isAnonymous}
\setboolean{isAnonymous}{false}  

\def\BibTeX{{\rm B\kern-.05em{\sc i\kern-.025em b}\kern-.08em
    T\kern-.1667em\lower.7ex\hbox{E}\kern-.125emX}}
\newcommand\f[1]{\textit{F1}(#1)}

\makeatletter
\newcommand{\newlineauthors}{%
  \end{@IEEEauthorhalign}\hfill\mbox{}\par
  \mbox{}\hfill\begin{@IEEEauthorhalign}
}
\makeatother

\begin{document}
\title{The Built-In Robustness of Decentralized Federated Averaging to Bad Data
\ifthenelse{\boolean{isAnonymous}}{}{\thanks{This work is partially supported by the European Union under the scheme HORIZON-INFRA-2021-DEV-02-01 – Preparatory phase of new ESFRI research infrastructure projects, Grant Agreement n.101079043, “SoBigData RI PPP: SoBigData RI Preparatory Phase Project”. 
This work was partially supported by SoBigData.it. SoBigData.it receives funding from European Union – NextGenerationEU – National Recovery and Resilience Plan (Piano Nazionale di Ripresa e Resilienza, PNRR) – Project: “SoBigData.it – Strengthening the Italian RI for Social Mining and Big Data Analytics” – Prot. IR0000013 – Avviso n. 3264 del 28/12/2021. 
S. Sabella's, C. Boldrini's and M. Conti's work was partly funded by the PNRR - M4C2 - Investimento 1.3, Partenariato Esteso PE00000013 - ``FAIR'', A. Passarella's and L. Valerio's work was partially supported by the European Union - Next Generation EU under the Italian National Recovery and Resilience Plan (NRRP), Mission 4, Component 2, Investment 1.3, CUP B53C22003970001, partnership on ``Telecommunications of the Future'' (PE00000001 - program “RESTART”).}
}
}

\ifthenelse{\boolean{isAnonymous}}{\author{\IEEEauthorblockN{Anonymous Authors}}}{
\author{\IEEEauthorblockN{Samuele Sabella}
\IEEEauthorblockA{\textit{Istituto di Informatica e Telematica} \\
\textit{Consiglio Nazionale delle Ricerche}\\
Pisa, Italy \\
samuele.sabella@iit.cnr.it}
\and
\IEEEauthorblockN{Chiara Boldrini\IEEEauthorrefmark{2}}
\IEEEauthorblockA{\textit{Istituto di Informatica e Telematica} \\
\textit{Consiglio Nazionale delle Ricerche}\\
Pisa, Italy \\
chiara.boldrini@iit.cnr.it}
\and
\IEEEauthorblockN{Lorenzo Valerio\IEEEauthorrefmark{2}}
\IEEEauthorblockA{\textit{Istituto di Informatica e Telematica} \\
\textit{Consiglio Nazionale delle Ricerche}\\
Pisa, Italy \\
lorenzo.valerio@iit.cnr.it}
\newlineauthors
\IEEEauthorblockN{Andrea Passarella}
\IEEEauthorblockA{\textit{Istituto di Informatica e Telematica} \\
\textit{Consiglio Nazionale delle Ricerche}\\
Pisa, Italy \\
andrea.passarella@iit.cnr.it}
\and
\IEEEauthorblockN{Marco Conti}
\IEEEauthorblockA{\textit{Istituto di Informatica e Telematica} \\
\textit{Consiglio Nazionale delle Ricerche}\\
Pisa, Italy \\
marco.conti@iit.cnr.it}
\thanks{\IEEEauthorrefmark{2} C. Boldrini and L. Valerio contributed equally to this work.}%
}
} 

\maketitle

\begin{abstract}
Decentralized federated learning (DFL) enables devices to collaboratively train models over complex network topologies without relying on a central controller. In this setting, local data remains private, but its quality and quantity can vary significantly across nodes. The extent to which a fully decentralized system is vulnerable to poor-quality or corrupted data remains unclear, but several factors could contribute to potential risks. Without a central authority, there can be no unified mechanism to detect or correct errors, and each node operates with a localized view of the data distribution, making it difficult for the node to assess whether its perspective aligns with the true distribution. Moreover, models trained on low-quality data can propagate through the network, amplifying errors. 
To explore the impact of low-quality data on DFL, we simulate two scenarios with degraded data quality—one where the corrupted data is evenly distributed in a subset of nodes and one where it is concentrated on a single node—using a decentralized implementation of FedAvg. Our results reveal that averaging-based decentralized learning is remarkably robust to localized bad data, even when the corrupted data resides in the most influential nodes of the network. Counterintuitively, this robustness is further enhanced when the corrupted data is concentrated on a single node, regardless of its centrality in the communication network topology. This phenomenon is explained by the averaging process, which ensures that no single node—however central—can disproportionately influence the overall learning process. 
\end{abstract}


\section{Introduction}
The Federated Learning paradigm leverages the use of multiple entities connected within a network, empowered with enough computational power to solve the task of knowledge extraction and model training. Each device extracts knowledge from locally available data by training a machine learning model and shares it with its neighbors, resulting in local knowledge spread across the network without the raw data leaving the device. Thus, problems that typically require access to all the data produced by the system, as done in centralized solutions, are solved using multiple learning devices trained in parallel on locally available data. Federated learning is studied within environments where it is not possible or convenient to share raw data. Healthcare facilities have strict privacy-related problems preventing them from sharing patients' data outside the infrastructure where they are collected or produced. Sharing locally trained neural networks has been one key solution to combine the knowledge coming from multiple sources and improve the predictions' performance~\cite{tedeschini2022decentralized}\cite{shiranthika2023decentralized}. Internet of Things (\textit{IoT}) networks usually have bandwidth limits that prevent sharing large volumes of data across the network. Sharing a compressed data representation may represent the only solution to learn a shared objective between devices using all the data available to the system. When computing power is not scarce and learning is required, decentralized learning could be a valid approach to avoid sending large quantities of data to a central server where learning is done via a centralized technique. Moreover, end users are increasingly concerned about privacy. Sharing only the knowledge extracted from the end devices without exposing the data used to extract that knowledge offers an effective alternative to centralized learning solutions that require all data to be accessible during training.

Compared to a centralized solution where data is centrally collected and trained in the cloud, federated learning introduces a greater number of degrees of freedom. The devices in the network can be initialized either homogeneously or heterogeneously, and data distribution across devices can be either identically and independently distributed (IID) or non-IID. In non-IID settings, each node possesses a unique perspective, differing from its neighbors. When devices learn exclusively from their local data, generalization across the entire system suffers, as no single node has access to the full data distribution. In star-like topologies, a central entity—typically called a Parameter Server (PS)—can coordinate knowledge sharing, a paradigm known as centrally coordinated federated learning (FL, hereafter). Conversely, when network conditions or topology prevent central coordination, the system operates under a fully decentralized federated learning paradigm (DFL, for short). In this paper, we focus on \emph{fully decentralized} learning systems with non-IID data distributions and homogeneous initialization, where no central coordination is available.

Previous research has demonstrated the robustness and resilience of decentralized learning to structural failures in network topology during training~\cite{palmieri2024robustness}. Specifically, DFL has shown that network disruptions have minimal impact on learning accuracy, provided that the remaining nodes retain sufficiently representative data. Moreover, most of the knowledge acquired before network disruptions is preserved even when some nodes become unreachable. Similarly, DFL can effectively handle highly skewed data distributions, as long as each node has access to a few representative examples~\cite{valerio2023coordination}. However, to the best of our knowledge, the impact of low-quality data on DFL has not been systematically investigated. Building on these insights, this paper examines the robustness of decentralized federated learning from a data quality perspective, focusing on the widely used yet simple model aggregation method based on averaging. We address the following research questions:
\begin{itemize}
\item \textbf{RQ1:} \emph{How sensitive is average-based decentralized federated learning to low-quality or corrupted data?}
\item \textbf{RQ2:} \emph{To what extent is this sensitivity influenced by the underlying network topology?}
\end{itemize}
To investigate these questions, we simulate decentralized environments where some nodes’ local datasets contain low-quality samples. In real-world scenarios, such data degradation may arise from sensor noise, faulty data augmentation, or intentional adversarial interference aimed at disrupting the learning process. Regardless of the cause, these misleading samples (referred to as \emph{corrupted data}) can affect DFL, and this paper aims to systematically assess their impact.

We generate malformed data using interpolations in the latent space of a pre-trained Generative Adversarial Network (\textit{GAN}), following the method described in~\cite{BKHMSILatentInterpolation}. This interpolation approach enables targeted corruption of specific labels with varying intensities, providing a controlled environment to analyze its effect on model performance.
We compare fully decentralized and federated learning approaches against a centralized baseline to assess their resilience to data corruption. Our experiments consider both balanced and imbalanced data distributions, including scenarios where central nodes (e.g., hubs) concentrate most of the available information. Additionally, we introduce increasing levels of feature corruption, ranging from a small fraction of the dataset up to 90\%, to evaluate the impact on learning outcomes.
 
Our key findings are the following:
\begin{itemize}
    \item Corrupted data primarily affects both the target and collateral classes, while overall accuracy remains largely stable, indicating minimal impact on other classes. Interestingly, corrupted samples from the target class that resemble the collateral class degrade the classifier’s performance more on the collateral class than on the target class itself.
    
    \item The distribution of bad data plays a crucial role in its impact: corruption spread across multiple nodes has a far more detrimental effect than when concentrated in a single node, regardless of that node’s influence in the network.
    
    \item When data corruption is unevenly distributed, decentralized learning proves to be more resilient than its centralized counterpart. While this does not definitively establish decentralized learning as the more robust approach, it highlights the need for further comparative analysis.
    
    \item Federated learning demonstrates superior long-term robustness to data corruption compared to a fully decentralized learning setup.
\end{itemize}

\section{Related Work}
We emphasize that this paper focuses on the susceptibility and robustness of decentralized learning to low-quality data distributed across the network topology. While this issue may bear some resemblance to adversarial machine learning, our perspective is fundamentally different, as we do not specifically address adversarial attacks. However, given the conceptual overlap, we include, alongside the relevant literature on decentralized learning, a brief overview of the literature on attacks against decentralized learning.

\subsection{Decentralized learning}
Most of the work studying the robustness of decentralized learning against data poisoning mainly focuses on attacks targeting federated learning and trying to compromise the model integrity by sending ad-hoc updates to the parameter server. Federated learning~\cite{mcmahan2017communication}, which currently has some of its applications in the field of user keyboard input prediction~\cite{hard2018federated} and healthcare~\cite{shiranthika2023decentralized}, is a form of decentralized learning where the underlying topology is star-like and the central server coordinates the learning. 

Decentralized learning, more broadly, offers a solution to learn within networks of devices, each holding a unique dataset, where constraints prevent the sharing of raw data across the network. The constraints are usually due to bandwidth limits or privacy requirements. Briefly, each device in the network starts by training a local model for a few epochs on the locally available data. Then, the distilled knowledge, usually the locally trained model, is shared with the neighbors. Finally, each device collects the models received by its neighbors and merges them into a new, updated model. Three steps are outlined: i) local training, ii) synchronization, iii) update. The three phases are then repeated until all devices reach convergence and models do not change anymore. It must be specified that local learning is analogous to the centralized paradigm, with all the techniques available such as early stopping on validation data extracted from the local dataset. Within these settings, learning is still possible even with non-IID data distributions and heterogeneous initialization~\cite{valerio2023coordination}. 

\subsection{Attacks against decentralized learning}

Unlike adversarial attacks that target models at test time~\cite{tabacof2016exploring}, our work focuses on the local training phase within a decentralized learning paradigm. Most prior research has concentrated on attacks against federated learning, along with corresponding countermeasures and detection mechanisms. In contrast, we aim to analyze overall network performance in a fully decentralized, uncoordinated, and potentially data-imbalanced scenario where nodes hold misleading and incorrectly labeled data.

Our scenario aligns closely with the category of \textit{label flip} attacks. In~\cite{bhagoji2019analyzing}, the authors investigated a federated IID setting with a single malicious user who manipulated gradient updates sent to the parameter server. Their approach leveraged either an additional scaling factor or a custom loss function to maximize the stealth and effectiveness of the attack. In~\cite{sun2024gan}, the authors proposed \textit{VagueGAN}, a modified GAN trained on an altered objective function that degrades the generator’s capabilities. In their experiments, they augmented the local dataset of each malicious node by 10\% using the pre-trained GAN and performed a label flip attack, converting all samples of class 6 to class 0 in the MNIST dataset. Their study examined scenarios with up to 30\% malicious participants.

Similarly, \cite{zhang2020poisongan} introduced \textit{PoisonGAN} to address cases where an attacker lacks access to a pre-trained generator or the necessary data distribution for training it. Their approach leveraged the global model managed by the parameter server as a discriminator to bypass data access limitations. Our work departs from these studies by assuming that nodes have full access to a pre-trained generator, which they use to construct or augment local datasets before training begins . 

In~\cite{tolpegin2020data}, the authors conducted an extensive study on label-flip poisoning in federated learning with IID data distributions. They examined the effects of poisoning by deploying multiple attackers at different time steps, specifically targeting selected labels. A comprehensive survey on poisoning attacks in federated learning is provided in~\cite{xia2023poisoning}. Meanwhile,~\cite{pham2024data} explored fully decentralized learning systems by injecting adversarial samples to induce backdoor behavior, forcing the victim model to respond to specific image triggers with predetermined outputs. This type of attack, known as a \textit{backdoor attack}, allows an adversary to embed a hidden objective in the model that can be exploited at test time. In contrast, our approach simply shifts the label of one class to another (specifically, class 4 to class 9) without introducing any additional objectives.

Unlike~\cite{gentz2015detection}, where malicious nodes execute adversarial code, our setting assumes that nodes have control only over their local dataset and do not run any harmful code. Lastly, we extend the ideas presented in~\cite{cao2019understanding}, where the impact of corrupted samples in federated scenarios was analyzed, to the extreme case of a fully decentralized setting with imbalanced data distribution. In our scenario, samples from two easily confusable classes are further corrupted using a generative model and assigned to the wrong class.

To the best of our knowledge, this is the first study to jointly investigate the effects of data corruption on a complex network topology in fully decentralized federated learning.

\begin{figure*}[t]
\includegraphics[width=\linewidth]{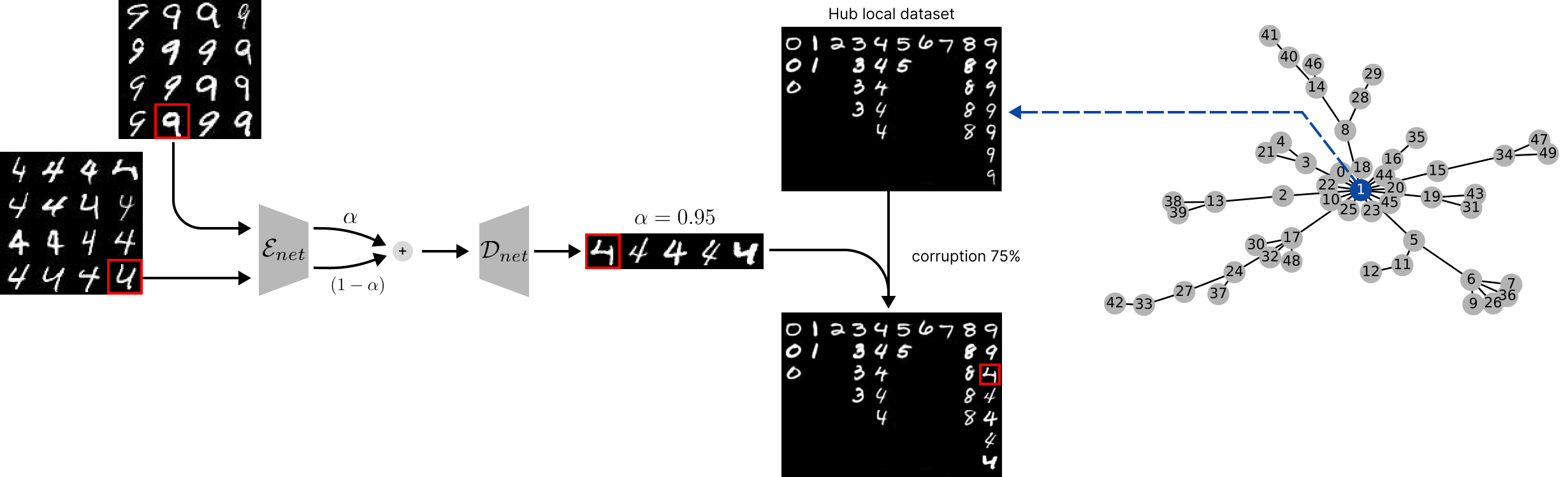}
\captionof{figure}{The hub's local dataset undergoes $75\%$ corruption with a corruption strength of $\alpha = 0.95$. The corrupted images are generated by interpolating samples from class four with samples from class nine using a pretrained GAN, where the interpolation is controlled by the parameter $\alpha$.}
\label{Fig:dataset_corruption}
\end{figure*}

\section{System Model}
\subsection{Decentralized learning}
\label{subsec:decentralized_learning}

We start by defining a network $\mathcal{G}(\mathcal{V},\mathcal{E})$ where $\mathcal{V}$ denotes the set of nodes and $\mathcal{E}$ the set of edges. Nodes represent individual learning entities (e.g., devices in a network), and edges represent the connections between them. We will assume that nodes represent devices in a network. Each device follows a three-step algorithm in every communication round: $i$) the device trains a local model using its locally available dataset for a few epochs; $ii$) local model's parameters are sent to the neighboring nodes; $iii$) the models received from the neighbourhood are combined through an \emph{aggregation function} and used to update the local model. A communication round consists of these three steps executed atomically. Decentralized learning proceeds through multiple communication rounds until the training converges across the network, i.e., the local model updates become negligible.

The overall dataset $\mathcal{D}$ can be denoted as the union of all local datasets: 
\[
\mathcal{D} \sim \mathcal{P} = \bigcup\limits_{i=1}^{N} D_i \sim  P_i \quad \text{with} \quad \bigcap\limits_{i=1}^{N} D_i = \emptyset,
\]
where \( D_i = \{(x_j, y_j)\}_{j=1}^{m} \) represents the local dataset of node \( i \), \( N \) is the total number of nodes in the network, and \( \mathcal{P} \) is the global data distribution. Since local data distribution may vary across devices, we explicitly define \( P_i \), the distribution from which \( D_i \) is drawn.

All devices share the same learning objective and underlying neural network architecture. However, local datasets may differ in size and label distribution. For instance, the hub (i.e., the most connected node in the network) may have a small, uniformly distributed dataset (\( \forall i: P_i \sim \mathcal{P} \)) or a large, imbalanced dataset. To isolate the effects of data corruption and avoid confounding factors introduced by model initialization, we assume homogeneous model initialization across devices.

Following~\cite{sun2022decentralized}, let \( h_i \) be the model of device \( i \) at communication round 0, parameterized by \( \mathbf{w}_i \). The device aims to minimize the local loss function \( \ell \) over its dataset \( D_i \):
\begin{equation}
    \tilde{\mathbf{w}}_i = \arg\min_{\mathbf{w}} \sum_{k=1}^{|D_i|} \ell(y_k, h(\mathbf{x}_k; \mathbf{w}_i)).
\end{equation}
At the end of the local training phase, the device \( i \) transmits \( \tilde{\mathbf{w}}_i \) to its neighboring nodes. The model for the next communication round is then updated by aggregating the received models in a layer-wise manner, weighted by the respective dataset sizes. Given \( \mathcal{N}(i) \), the set containing node \( i \) and its neighbors, the model update at communication round \( t \) is defined as:
\begin{equation} \label{eq:aggr}
    \mathbf{w}_i^{(t)} \leftarrow
    \frac{\sum_{j \in \mathcal{N}(i)} |D_j| \tilde{\mathbf{w}}_j^{(t-1)}}
    {\sum_{j \in \mathcal{N}(i)} |D_j|}.
\end{equation}
Once \( \mathbf{w}_i^{(t)} \) is updated, the local learning process restarts, and the cycle repeats until convergence.
The described approach aligns and generalizes with respect to previous work~\cite{palmieri2024robustness,valerio2023coordination,sun2022decentralized,savazzi2020federated}. The aggregation in Eq.~\eqref{eq:aggr} performs a weighted averaging of the model parameters, akin to the widely used FedAvg\cite{mcmahan2017communication} approach (proposed for centralized FL), but adapted to a decentralized setting.

\subsection{Dataset corruption}
We extend the notation introduced in the previous section by defining $C(\cdot; \mathcal{I}, s)$, a function that maps $\mathcal{D}$ to $\tilde{\mathcal{D}}$, the corrupted dataset. In our case, $\tilde{\mathcal{D}}$ is just $\mathcal{D}$ being subject to a post-processing function $\mathcal{I}$ which targets a specific label class $c_t$ (\emph{target class}), thus:
\begin{equation}
\mathcal{C}(\mathcal{D}; \mathcal{I}, s) = \tilde{\mathcal{D}} =
  \begin{cases}
  (\mathcal{I}(x_i), y_i) & \text{if } \ \ y_i = c_t, \\
  (x_i, y_i) & \text{otherwise}.
  \end{cases}
\end{equation}

To analyze the network's behavior under varying levels of corruption, we conducted experiments where samples for label $c_t$ were drawn either from $\mathcal{D}$ or $\tilde{\mathcal{D}}$ with percentage $p$ gradually increasing from $10\%$ to $90\%$. Given $\mathcal{D}=\{x_i, y_i\}_{1..n}$ and $\tilde{\mathcal{D}}=\{\tilde{x}_i, y_i\}_{1..n}$ we define our final dataset, from which local datasets are drawn, as:
\begin{equation}
\tilde{\mathcal{D}}_p = \{(x_i, y_i)\}_{1..\lfloor p*n \rfloor} \cup \{(\tilde{x}_i, y_i)\}_{\lceil p*n \rceil ..n}
\end{equation}
Our downstream task is supervised image classification. Within this context, we pick two classes, $c_c$ and $c_t$, and we corrupt the features of $c_t$ with an interpolation towards class $c_c$. This corruption makes the learning model more prone to errors when discriminating between the two classes~\cite{sikar2024misclassification}. Interpolated images are built upon the latent representation of a pre-trained GAN~\cite{goodfellow2014generative}, a special case of artificial curiosity~\cite{schmidhuber2020generative}\cite{schmidhuber1990making}. Let   $\mathcal{E}_{net}$ and $\mathcal{D}_{net}$ be the encoder and decoder of our model trained using an adversarial loss, respectively. For each sample $x_i$ subject to corruption, let $x_c, x_t$ be samples randomly drawn from the collateral class $c$ and target class $t$, respectively. We obtain the corrupted sample $\tilde{x}_i$ by combining their latent representation via a simple linear transformation with parameter $\alpha \in [0,1]$ controlling the interpolation strength. We then used the interpolated sample in place of $x_i$.
\begin{equation} \label{eq:corruption}
\tilde{x}_i=\mathcal{I}(x_i) = \mathcal{D}_{net} (\alpha * \mathcal{E}_{net}(x_t) + (1-\alpha) * \mathcal{E}_{net}(x_c))
\end{equation}
Figure~\ref{Fig:dataset_corruption} summarizes the process we used to corrupt the datasets in our experiments, applied to the MNIST dataset, where $x_c$ belongs to class 4 and $x_t$ belongs to class 9. In the example, the interpolation transforms the features of class 9 towards those of class 4.   

\section{Experimental Settings}

\subsection{Communication network}

We run experiments for scenarios composed of $50$ nodes both for the federated learning and the fully decentralized setting. In FL the topology is constrained to being star-like (with the coordinating server at the center). In our federated learning setup, we assume that all devices participate in every communication round, whereas some FL algorithms allow only a fraction of nodes to contribute at each round. This choice removes one degree of variability in our experiments, ensuring a consistent participation pattern across all nodes. 

In the decentralised scenarios, we employ a Barabási–Albert (BA) graph topology~\cite{barabasi1999emergence}, which belongs to the family of scale-free networks. These networks are characterized by a few highly connected nodes, known as hubs, which play a crucial role in information spreading due to their extensive connectivity. Given their central position, hubs serve as our primary targets for the investigation, as their models are directly shared with numerous other devices. We generate BA graphs using the NetworkX Python library~\cite{hagberg2008exploring}, setting the parameter  $m$ to $1$, meaning that each newly added node connects to only one existing node. Because the BA model preferentially attaches new nodes to those with higher degrees, this results in a small number of well-connected hubs, while most nodes remain sparsely connected, often with just one link (because $m=1$). 

In both FL and DFL, nodes communicate exclusively with their neighbors in the network. However, in FL, each node has a single neighbor (the central server), whereas in DFL, nodes have varying numbers of neighbors depending on the degree distribution of the communication graph. The BA topology is particularly interesting for studying DFL because its heavy-tailed degree distribution (a common feature in many real-life networks, including social and technological systems~\cite{barabasi2013network}) introduces maximum heterogeneity in node connectivity and communication opportunities.

\subsection{Dataset and data distribution}
\label{sec:data}

\begin{figure}[!t]
    \centering
    \includegraphics[width=0.5\linewidth]{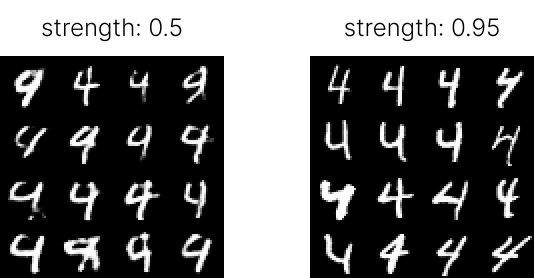}
    \caption{Examples of interpolated images of nine with images of class four with $\alpha=0.5$ (left) and $\alpha=0.95$ (right).}
    \label{fig:interpolation-strength}
\end{figure}

In our experiments we use the MNIST dataset~\cite{lecun1998mnist}. Given its simplicity and well-defined classes, it provides an ideal testbed for analyzing and understanding the impact of bad data in a controlled manner.
The MNIST dataset consists of $60,000$ training samples of handwritten digits from $0$ to $9$ and a test set containing $10,000$ samples. Each sample consists of a $28 \times 28$ image and a label reporting the digit number. The local datasets of each node are mutually exclusive partitions of the training set, the test set is common to all the nodes. We investigated a scenario where low-quality data affects a specific class. To this aim, we perform label interpolation modifying images labeled as $9$ to resemble those labeled as $4$. We explore interpolation strength (i.e. $\alpha$ in eq.~\ref{eq:corruption}) set to $0.5$ and to $0.95$, the latter basically resulting in a naive label flip. Figure~\ref{fig:interpolation-strength} shows some examples of interpolated samples at different strength levels. 
We refer to class $9$ as the \textbf{target class}, as its samples are directly modified, and to class $4$ as the \textbf{collateral class}, since the increasing resemblance of corrupted $9$s to $4$s may indirectly distort the learning process for class $4$. The remaining classes are referred to as \textbf{bystanders or uninvolved classes}. While the parameter $\alpha$ controls the extent to which corrupted $9$s resemble $4$s, the actual number of corrupted samples in class $9$ is governed by parameter $p\in[0,1]$. Consequently, the impact of poor-quality data is maximized when both $\alpha$ and $p$ are close to $1$, i.e., the totality of the target class sample is indistinguishable from those of the collateral class, but the labels remain unchanged, resulting in a sort of label flip.

In our experiments, images from all classes except class $9$ are assigned in an IID manner, ensuring uniform distribution across all nodes. This approach prevents unintended effects where some nodes might receive too little data, which could degrade performance for reasons unrelated to the presence of bad data. By maintaining a balanced data allocation for uninvolved and collateral classes, we ensure that the observed impact is primarily attributable to the presence of corrupted samples.

For class $9$, we consider the following distributions:

\noindent\textbf{Balanced bad data distribution}: Here, the fraction $p$ of corrupted samples is evenly distributed among the most central nodes, which, as previously discussed, play a more influential role in the communication network. Specifically, the $\lceil pN \rceil$ most central nodes receive the corrupted samples\footnote{If the total number of training samples for class \( 9 \) is \( n_9 \), each node is expected to receive approximately \( \frac{n_9}{N} \) samples under a uniform distribution. By concentrating the corrupted samples on the most central nodes, the total number of corrupted samples, \( p n_9 \), is allocated in groups of \( \frac{n_9}{N} \). Thus, the number of nodes affected by corruption is given by \( \frac{p n_9}{n_9/N} = pN \), meaning that \( \lceil pN \rceil\) is the number of nodes receiving corrupted samples.}, meaning that within these nodes, all instances of class $9$ are replaced with corrupted samples.
We allocate bad samples in descending order of centrality, ensuring that only the most central nodes receive them. This setting is termed \textit{balanced} because all affected nodes hold the same number of bad samples, with minor variations due to border effects (e.g., when the number of bad samples is not an integer multiple of the local dataset size for class $9$).
    
\begin{figure}[t!]
    \centering
    \includegraphics[width=\linewidth]{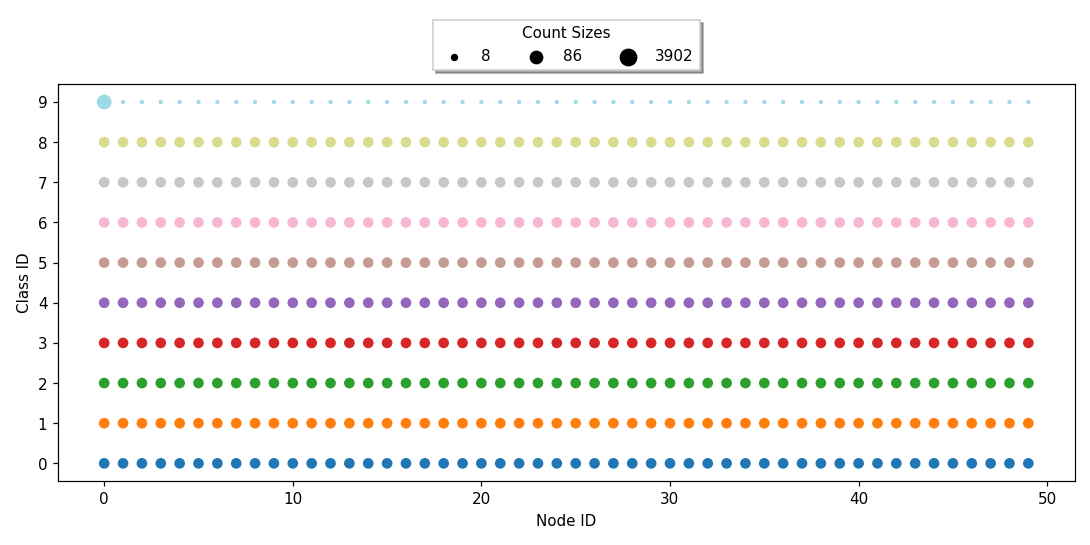}
    \caption{Example of unbalanced corrupted data distribution.}
    \label{fig:nonIID_data_scatter_plot}
\end{figure}

\noindent\textbf{Unbalanced bad data distribution}: In this scenario, all corrupted samples are assigned to the most central node, while the remaining nodes receive an equal share of the unaltered samples. Thus, while in the balanced bad data scenario $\lceil pN \rceil$ central nodes receive low-quality data, here only one node gets \textit{all} the corrupted data. Figure~\ref{fig:nonIID_data_scatter_plot} provides a visual representation of this distribution, when the hub holds $90\%$ of class-9 samples.

\begin{figure}[t!]
    \centering
    \includegraphics[width=\linewidth]{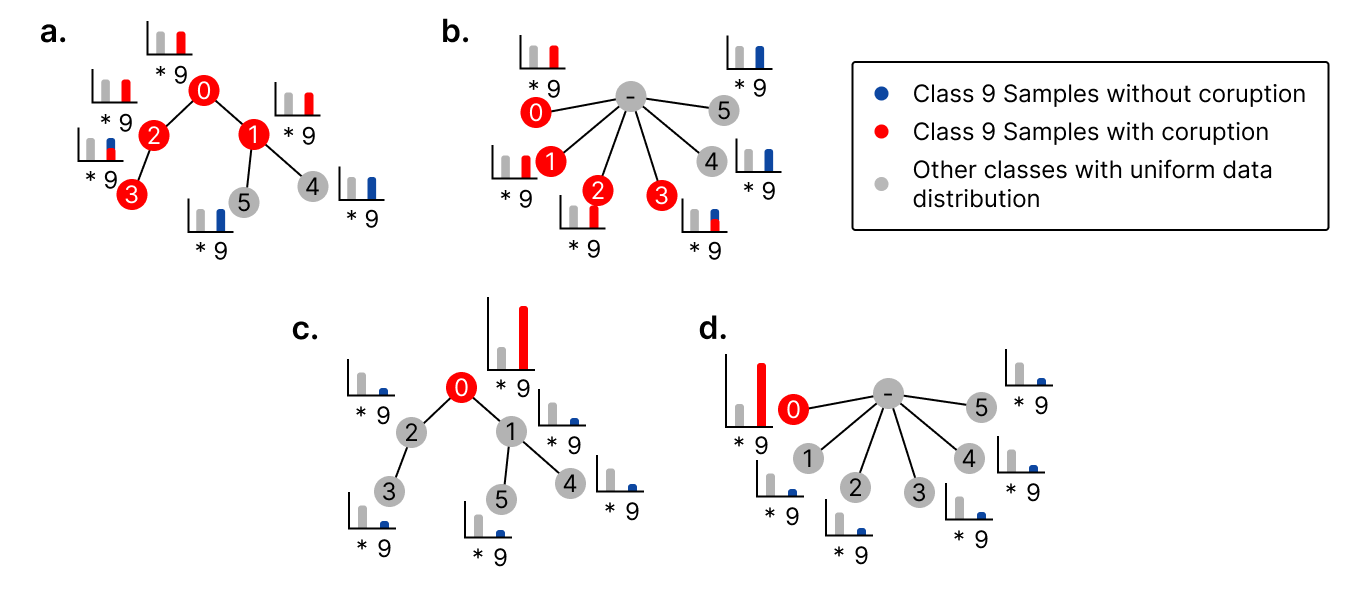}
    \caption{We simulated four main scenarios: (a) decentralized with balanced bad data distribution, (b) federated with balanced bad data distribution, (c) decentralized with unbalanced bad data distribution, and (d) federated with unbalanced bad data distribution. Non-target classes (i.e., all classes except class $9$) are distributed uniformly.}
    \label{fig:data_distribution}
\end{figure}

Table~\ref{tab:datadist} summarizes the data distribution under the two scenarios. Figure~\ref{fig:data_distribution} illustrates the different configurations tested in our simulations. Note that we mimic the balanced and unbalanced corrupted data distributions when testing FL. However, since FL is inherently constrained to a star-like topology, where each device has exactly one connection (to the central server), node centrality does not influence the learning process in this setting.

\begin{table}[!t]
\renewcommand{\arraystretch}{2}
\centering
\caption{Data distribution summary. $v_{(i)}$ denotes the $i$-th most central node. $n_i$ is the number of training samples in class $i$. $N$ is the number of nodes. $p$ is the fraction of corrupt samples.}
\label{tab:datadist}
\begin{tabular}{@{}cccccc@{}}
\toprule
\textbf{} & \textbf{Class $i \in \{0,8\}$} & \multicolumn{2}{c}{\textbf{Bal. Class 9}} & \multicolumn{2}{c}{\textbf{Unbal. Class 9}} \\ \cmidrule(lr){3-4} \cmidrule(lr){5-6} 
             &                 & \textbf{Good}   & \textbf{Bad}    & \textbf{Good}           & \textbf{Bad} \\ \midrule
$v_{(1)}$    & $\dfrac{n_i}{N}$ & 0               & $\dfrac{n_9}{N}$ & 0                       & $p n_9$      \\
...          & $\dfrac{n_i}{N}$ & 0               & $\dfrac{n_9}{N}$ & $\dfrac{(1-p) n_9}{N-1}$ & 0            \\
$v_{(pN)}$   & $\dfrac{n_i}{N}$ & 0               & $\dfrac{n_9}{N}$ & $\dfrac{(1-p) n_9}{N-1}$ & 0            \\
$v_{(pN+1)}$ & $\dfrac{n_i}{N}$ & $\dfrac{n_9}{N}$ & 0               & $\dfrac{(1-p) n_9}{N-1}$ & 0            \\
...          & $\dfrac{n_i}{N}$ & $\dfrac{n_9}{N}$ & 0               & $\dfrac{(1-p) n_9}{N-1}$ & 0            \\
$v_{(N)}$    & $\dfrac{n_i}{N}$ & $\dfrac{n_9}{N}$ & 0               & $\dfrac{(1-p) n_9}{N-1}$ & 0            \\ \bottomrule
\end{tabular}
\end{table}

\subsection{Learning \& simulation settings}

Each node locally trains a CNN configured according to state-of-the-art practices with a cross-entropy supervised learning objective. The CNN consists of two convolution layers (kernel size is set to $5$), each followed by max pooling and a ReLu activation function. Dropout interleaves the two layers and the output is given by two fully connected layers with a ReLu activation function and dropout applied between the two. We used the stochastic gradient descent with learning rate set to $1e-3$ and momentum set to $0.9$. In all our experiments we set the batch size to $32$. We also set the local validation size to $20\%$ of all the training set local to each node. We run simulations up to $1,000$ communication rounds, with each node training for a maximum of $5$ local epochs. Early stopping is applied to prevent local model degradation. We replicate our simulations varying both the seed governing graph generation and the seed controlling other sources of randomness beyond the network topology. We also present the results for a centralized benchmark, where the same CNN is trained on the overall dataset $\mathcal{D}$.

\subsection{Performance metrics}
We evaluated each node’s performance using a common test set shared across all devices. At each communication round, we collected the confusion matrix for each node and extracted two key metrics: \textit{accuracy} and \textit{F1 score}. Depending on the analysis, we present these metrics in different ways: averaged across all nodes, within the neighborhood of target nodes, or for a specific subset of nodes. We will specify the chosen approach as needed throughout the discussion. When not specified, the results we present in this paper report the mean value at each communication round together with the $95\%$ confidence interval computed over the different seeds used.



\section{Results}

Before proceeding with a detailed analysis, we first highlight a key observation: our experiments indicate that decentralized federated averaging is largely robust to the presence of corrupted data in the system. Specifically, we found that data corruption effects become clearly pronounced when its strength is set to $\alpha=0.95$. Consequently, unless stated otherwise, all results presented in this work have been obtained using this corruption strength. Refer to Figure~\ref{fig:interpolation-strength} for a qualitative illustration of the impact of corruption for different $\alpha$.

We begin by illustrating the impact of low-quality data over time in Figure~\ref{fig:dec,iid,corruption_strength.95}, where we vary the dataset corruption percentage (\( p \) in \( \tilde{\mathcal{D}}_p \)). The figure presents results for DFL in the most vulnerable scenario identified: balanced corrupted data distribution.
The plot tracks performance at each communication round using three key metrics. The first is \textit{accuracy}, which reflects the overall impact of the attack on all classes. The second is the \textit{F1 score for class 9}, denoted as \f{9}, quantifying degradation in the corrupted class. The third is the \textit{F1 score for class 4}, denoted as \f{4}, capturing potential misclassification side effects in the collateral class.  
With no corrupted data, the performance of decentralized learning soon approaches that of a centralized approach. However, as $p$ increases, the impact of low-quality data for class $9$ is clearly visible, and as expected, its severity increases with the corruption fraction \( p \). As the proportion of bad samples grows, accuracy and F1 scores progressively decline. However, the overall accuracy experiences only a mild drop, with at most a 10 percentage point decrease compared to an uncorrupted scenario. Notably, the corruption of class 9 strongly affects class 4, leading to increasing misclassification, i.e., low \f{4}. \textit{These results suggest that low-quality data primarily affects both the target class (class 9) directly and the collateral class (class 4) indirectly, while having minimal impact on the remaining classes.}

\begin{figure}[t!]
    \centering
    \includegraphics[width=\linewidth]{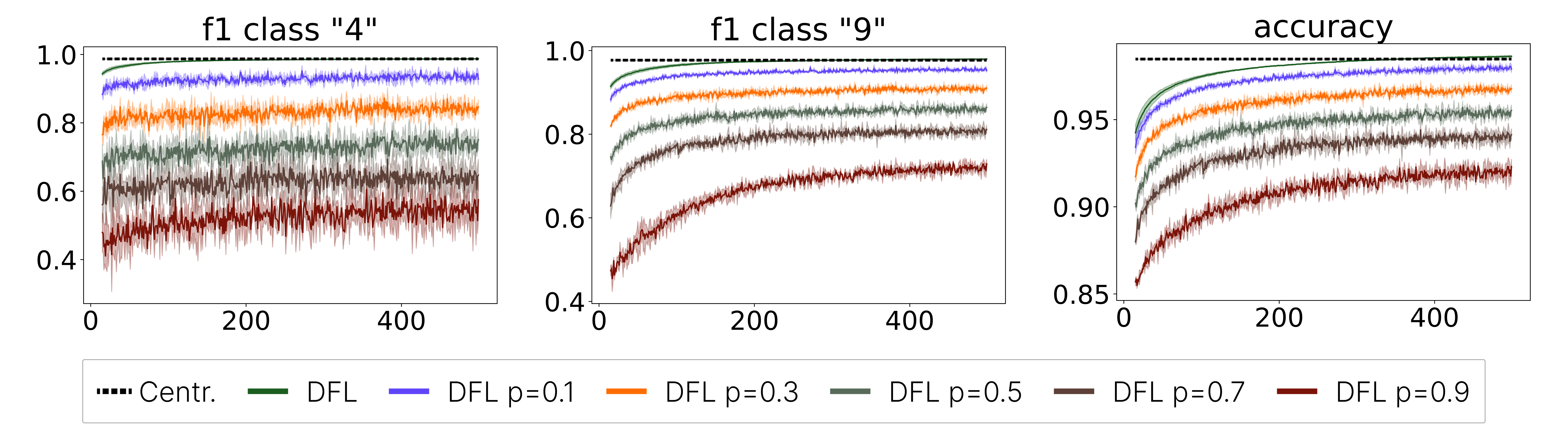}
    \caption{Comparison of the performance between a centralized paradigm (dashed line) and DFL (continuous line) under \textbf{balanced corrupted data distribution}, with a corruption strength of $\alpha=95\%$ and fraction of bad samples ranging from $p=0.1$ up to $p=0.9$.}
    \label{fig:dec,iid,corruption_strength.95}
\end{figure}

Table~\ref{tab:centralized_baseline} presents the F1 score for the best-performing round under high corruption conditions, i.e., with a corruption strength of $\alpha=95\%$ and $p=90\%$ of class 9 samples globally corrupted. For comparison, the table also includes baseline performance in the absence of corruption. A key observation is that, in a corruption-free scenario, FL, DFL, and centralized learning achieve rather similar performance (highlighted in grey in Table~\ref{tab:centralized_baseline}). However, when corrupted data is introduced, all learning strategies experience performance degradation, as indicated by the green-highlighted cells showing lower F1 scores. Interestingly, the impact of corruption varies depending on the distribution of corrupted data across nodes. In the unbalanced case (highlighted in light green), the performance drop remains relatively small, with a worst-case loss of 8 percentage points (pp) with respect to the corresponding configuration without corruption. In contrast, in the balanced corruption scenario (darker green), the degradation is much more severe, with class 4 (the collateral class) losing approximately 30 pp and class 9 (the target class) losing around 20 pp.
\textit{Surprisingly, the performance drop is more severe for class 4 than for class 9. This suggests that classifying 4s becomes substantially more challenging due to the presence of corrupted samples labeled as 9s that visually resemble 4s. Conversely, corruption on 9s appears to act—counterintuitively—as a form of generalization, making it easier for the model to handle class 9. }As a result, while the model does make mistakes on 9s, it makes significantly more errors on 4s, whose dataset remains unaltered yet is indirectly affected by the misleading samples introduced into class 9.

\definecolor{Balanced}{RGB}{140, 210, 140} 
\definecolor{Unbalanced}{RGB}{195, 255, 180} 

\begin{table}[!t]
\renewcommand{\arraystretch}{1.3}
\caption{Comparison of final F1 scores w/o corruption and w/ corruption ($\alpha=0.95$ and $p=0.9$).}
\label{tab:centralized_baseline}
\centering
\begin{tabular}{ccccc}
\toprule
\textbf{Corr.?}   & \textbf{Method}       & \textbf{Bad Data} & \textbf{F1(4)}                & \textbf{F1(9)}                \\ 
 & &\textbf{Dist.}& &  \\\midrule
                     & Centr.           & \cellcolor{lightgray}-          & \cellcolor{lightgray}$0.991\pm0.000$                         & \cellcolor{lightgray}$0.980\pm0.000$                          \\ \cmidrule(l){2-5} 
                     &                       & \cellcolor{lightgray}Balanced   & \cellcolor{lightgray}$0.992\pm0.000$                         & \cellcolor{lightgray}$0.985\pm0.000$                         \\
                     & \multirow{-2}{*}{FL}  & \cellcolor{lightgray}Unbalanced & \cellcolor{lightgray}$0.986\pm0.002$                         & \cellcolor{lightgray}$0.976\pm0.001$                         \\ \cmidrule(l){2-5} 
                     &                       & \cellcolor{lightgray}Balanced   & \cellcolor{lightgray}$0.989\pm0.001$                         & \cellcolor{lightgray}$0.982\pm0.002$                         \\
\multirow{-5}{*}{No} & \multirow{-2}{*}{DFL} & \cellcolor{lightgray}Unbalanced & \cellcolor{lightgray}$0.986\pm0.002$                         & \cellcolor{lightgray}$0.976\pm0.002$                         \\ \midrule
                     & Centr.           & \cellcolor{Unbalanced}-          & \cellcolor{Unbalanced}$0.837 \pm 0.000$ & \cellcolor{Unbalanced}$0.831 \pm 0.000$ \\ \cmidrule(l){2-5} 
                     &                       & \cellcolor{Balanced}Balanced   & \cellcolor{Balanced}$0.666 \pm 0.000$ & \cellcolor{Balanced}$0.790 \pm 0.000$  \\
                      & \multirow{-2}{*}{FL}  & \cellcolor{Unbalanced}Unbalanced              & \cellcolor{Unbalanced}$0.952\pm0.001$ & \cellcolor{Unbalanced}$0.929\pm0.003$ \\ \cmidrule(l){2-5} 
                     &                       & \cellcolor{Balanced}Balanced   & \cellcolor{Balanced}$0.640\pm0.034$ & \cellcolor{Balanced}$0.758\pm0.011$ \\
\multirow{-5}{*}{Yes} & \multirow{-2}{*}{DFL} & \cellcolor{Unbalanced}Unbalanced              & \cellcolor{Unbalanced}$0.927\pm0.007$ & \cellcolor{Unbalanced}$0.886\pm0.014$ \\ \bottomrule
\end{tabular}
\vspace{0.5em} 
\vspace*{-\baselineskip} 
\end{table}

Figures~\ref{fig:iid,corruption_strength.95} and~\ref{fig:noniid,corruption_strength.95} report accuracy, \f{4}, \f{9} for both FL and DFL. In the \emph{balanced bad data} configuration (Fig.~\ref{fig:iid,corruption_strength.95}), we observe curves that are clearly separated for different values of $p$. From the temporal evolution standpoint, the figure highlights that FL also converges faster than DFL over time, especially for small $p$. In the unbalanced bad data scenario (Fig.~\ref{fig:noniid,corruption_strength.95}), where corrupted samples are concentrated on a single node, performance differences are minimal for low to intermediate values of  $p$, with FL and DFL exhibiting nearly identical behavior. For high  $p$, DFL initially learns faster but is eventually overtaken by FL, which surpasses its performance at later stages. As previously noted, the unbalanced bad data case has a significantly smaller impact on overall performance, as the corruption remains localized and does not propagate widely across the network.


\begin{figure}[!t]
    \centering
    \includegraphics[width=\linewidth]{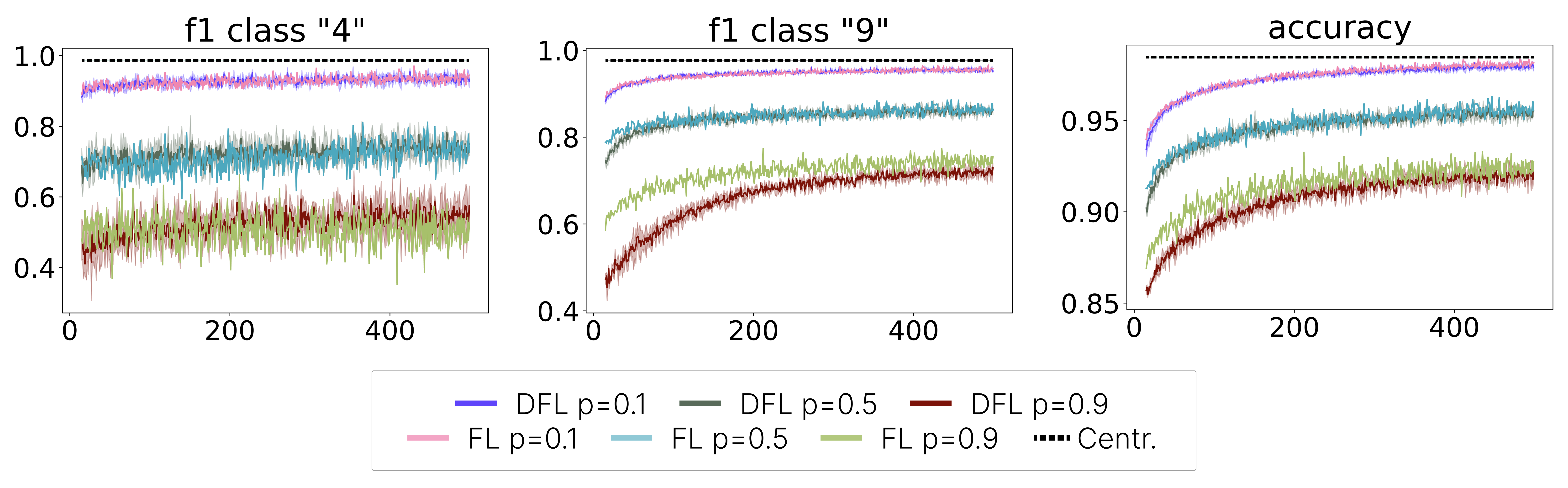}
    \caption{Balanced corrupted data: DFL vs FL, with corruption strength $95\%$.}
    \label{fig:iid,corruption_strength.95}
\end{figure}
\begin{figure}[!t]
    \centering
    \includegraphics[width=\linewidth]{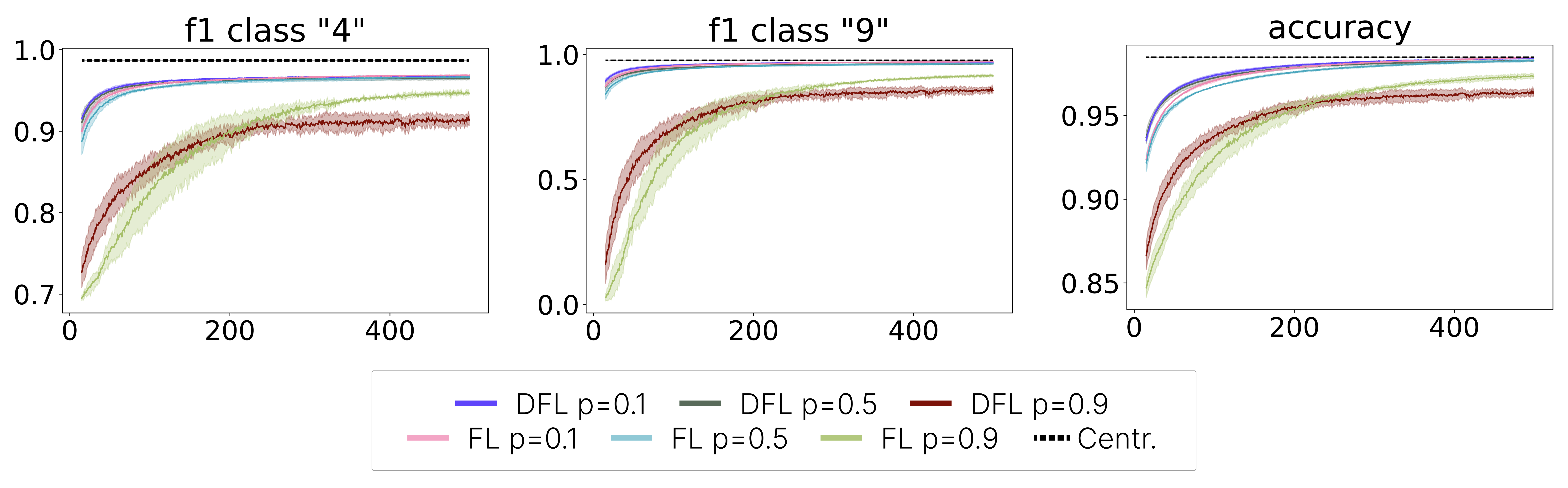}
    \caption{Unbalanced corrupted data: DFL vs FL, with corruption strength $95\%$.}
    \label{fig:noniid,corruption_strength.95}
\end{figure}

\begin{figure*}[!t]
    \centering
    \includegraphics[width=\linewidth]{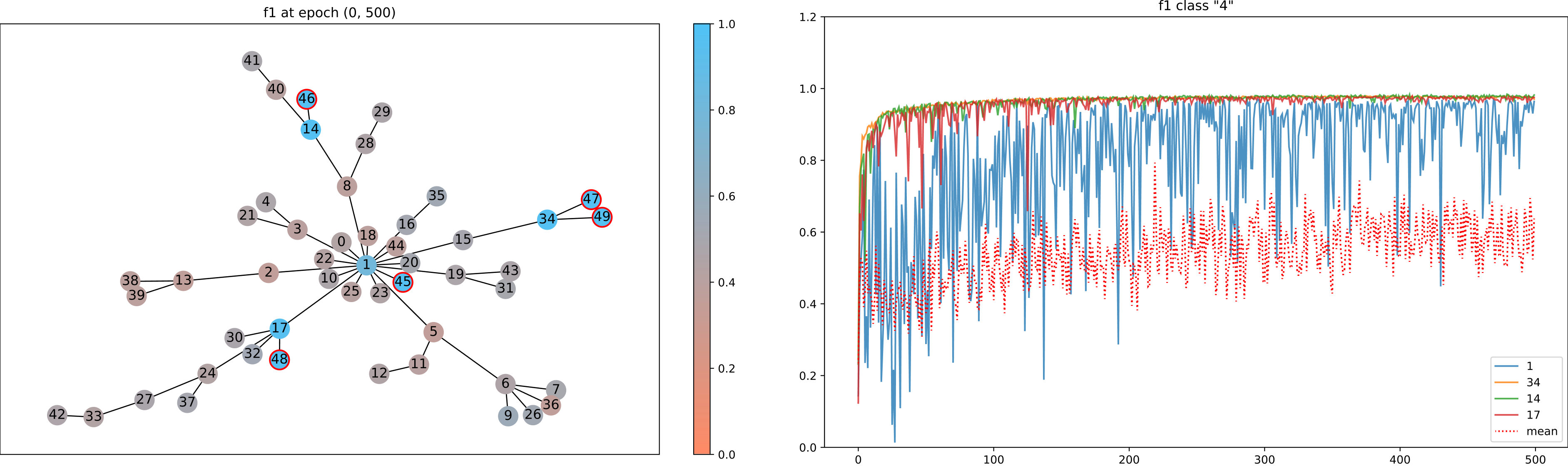}
    \caption{Example of a fully decentralized learning system with $\alpha=.95$ and $p=0.9$. Left: the network topology for one of our experiments. The color of the nodes represents the mean \f{4} score computed after the aggregation phase for class $4$ after $500$ communication rounds. The nodes with a red circle are the ones not containing corrupted samples. Right: Mean F1 score computed after the aggregation phase for class $4$ over communication rounds. The red line reports the mean \f{4} score computed across all the nodes at each communication round.}
    \label{fig:topology-F1-4}
\end{figure*}

Taken together, these results indicate that, for the same amount of bad-quality data,\textit{ the impact is significantly greater when the corrupted samples are distributed across multiple nodes rather than concentrated within a single node, regardless of the node’s centrality.}
%
Building on this finding, we further argue that having a neighbor with uncorrupted data helps mitigate the effects of corruption. Figure~\ref{fig:topology-F1-4} presents an analysis of a DFL simulation with $p = 0.9$. The leftmost figure visualizes the network topology, where each node is colored based on its mean \f{4}. Nodes circled in red represent devices that do not contain corrupted samples. The rightmost figure plots the mean \f{4} score across the entire network against the \f{4} scores of individual nodes, particularly those connected to nodes with uncorrupted data. 
The results clearly show that having at least one ``clean" neighbor significantly mitigates the negative effects of corrupted data, leading to a local \f{4} score that exceeds the network-wide average. From this, we conclude that denser networks may exhibit greater resilience to low-quality data when corruption is confined to a subset of nodes. In such networks, uncorrupted nodes have more connections, potentially acting as stabilizing points in the aggregation process and mitigating the impact of corrupted data.



The increased robustness of federated learning compared to a fully decentralized approach becomes less noticeable in milder scenarios. By using a corruption strength of $\alpha=0.5$ both presented similar convergence for the balanced and unbalanced scenario as shown in Figures~\ref{fig:iid,corruption_strength.5} and~\ref{fig:noniid,corruption_strength.5}.

\begin{figure}[!t]
    \centering
    \includegraphics[width=\linewidth]{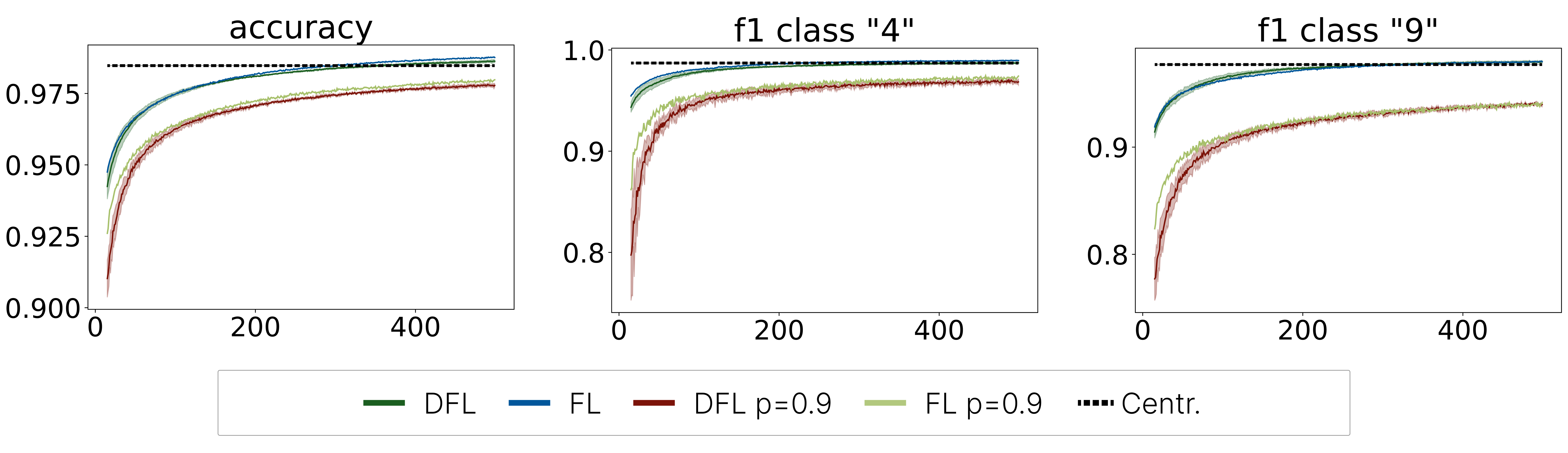}
    \caption{Corruption strength $\alpha=0.5$ with balanced corrupted data: DFL vs FL for varying $p$.}
    \label{fig:iid,corruption_strength.5}
\end{figure}
\begin{figure}[!t]
    \centering
    \includegraphics[width=\linewidth]{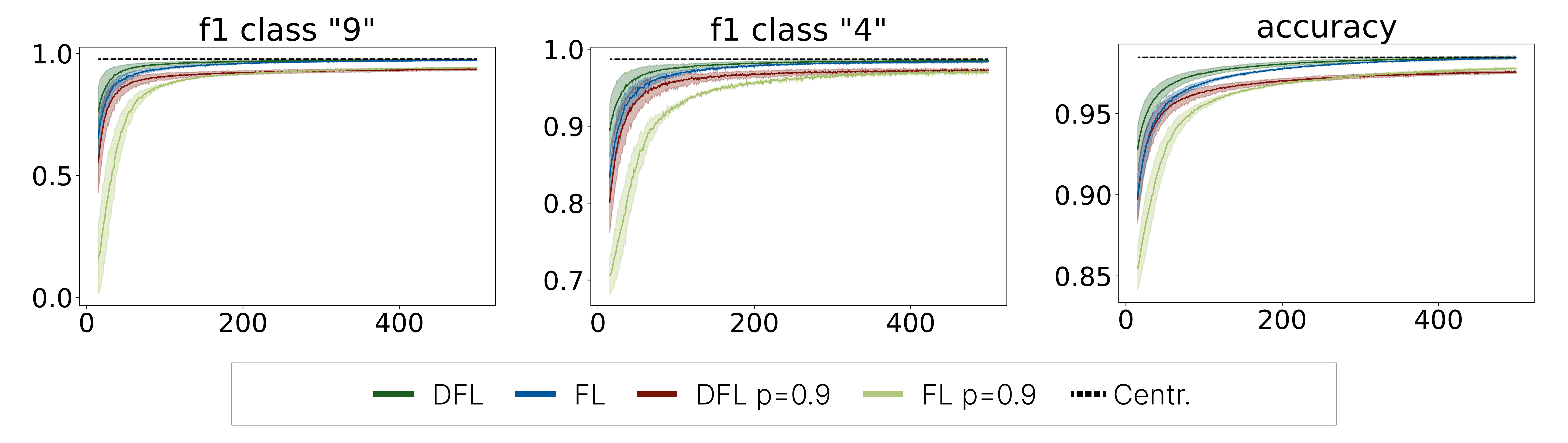}
    \caption{Corruption strength $\alpha=0.5$ with unbalanced corrupted data: DFL vs FL for varying $p$.}
    \label{fig:noniid,corruption_strength.5}
\end{figure}

To summarize, the following findings emerged from our observations:
\begin{itemize}
    \item The impact of corrupted data is most pronounced in the F1 scores for class 4 and class 9, while overall accuracy is only mildly affected, suggesting that other classes remain largely unaffected. Surprisingly, corrupted 9s that resemble 4s degrade the performance of the decentralized classifier more on class 4 than on class 9.
    \item The same amount of bad data has a significantly greater impact when distributed across multiple nodes rather than concentrated in a single, even highly influential, node.
    \item Decentralized learning under an unbalanced data distribution outperforms its centralized counterpart. While this does not necessarily imply that decentralized learning is inherently more robust, it warrants further investigation.
    \item Federated learning exhibits greater long-term resilience to corrupted data compared to a fully decentralized approach.
\end{itemize} 

\section{Discussion and Conclusion}
In this work, we investigated the impact of corrupted data on decentralized federated learning (DFL) and compared its resilience to federated learning (FL) and centralized learning. Our evaluation was conducted on a realistic communication network modeled as a heavy-tailed Barabási–Albert (BA) graph, with a focus on a standard image classification task. We simulated decentralized environments where some nodes’ local datasets contained low-quality samples, mimicking real-world scenarios in which data degradation arises from sensor noise, faulty data augmentation, or intentional adversarial interference.  
To introduce controlled corruption, we leveraged interpolations in the latent space of a pre-trained GAN, enabling targeted corruption of specific labels with varying intensities. Our experiments examined both balanced and unbalanced corruption distributions, considering scenarios where corrupted data was either spread across multiple nodes or concentrated on a single node, including highly connected hubs. We also explored different corruption levels, ranging from minor perturbations to extreme cases where up to 90\% of class 9 samples were corrupted.  

Our key findings indicate that corruption primarily affects the target and collateral classes while leaving other classes largely unaffected. Interestingly, misclassified target samples that resemble the collateral class degrade performance more on the collateral class than on the target class itself. Furthermore, we observed that the distribution of corruption plays a crucial role: when corrupted data is scattered across multiple nodes, the performance drop is significantly larger than when it is concentrated in a single node, even if that node is highly central. This suggests that having uncorrupted neighbors helps mitigate the negative effects of bad data.  
Additionally, we found that under an unbalanced data distribution, DFL can outperform a naive centralized approach in handling corruption. However, this does not necessarily indicate that decentralized learning is universally more robust; rather, its resilience depends on the corruption pattern and network structure, warranting further investigation. Lastly, while fully decentralized learning initially converges faster, FL exhibits greater long-term robustness, ultimately surpassing DFL in later communication rounds.  

Our findings provide new insights into the resilience of decentralized learning systems to low-quality data, emphasizing the strong interplay between network structure and the distribution of corrupted data. This interplay suggests potential design strategies for peer-to-peer overlays that enhance collaboration among nodes when network constraints allow. Future research should explore alternative aggregation strategies beyond simple averaging, extend the analysis to different network topologies, and evaluate additional datasets.  

Finally, given recent studies on the decentralized training of diffusion models to alleviate the computational burden of training infrastructure~\cite{mcallister2025decentralized}, we argue that a deeper understanding of how unbalanced settings impact performance could inform more effective data distribution strategies. This, in turn, may help mitigate the risks associated with training on unverified data sources, such as those scraped from uncontrolled internet environments.  

\bibliographystyle{IEEEtran}
\bibliography{references}

\begin{thebibliography}{10}
\providecommand{\url}[1]{#1}
\csname url@samestyle\endcsname
\providecommand{\newblock}{\relax}
\providecommand{\bibinfo}[2]{#2}
\providecommand{\BIBentrySTDinterwordspacing}{\spaceskip=0pt\relax}
\providecommand{\BIBentryALTinterwordstretchfactor}{4}
\providecommand{\BIBentryALTinterwordspacing}{\spaceskip=\fontdimen2\font plus
\BIBentryALTinterwordstretchfactor\fontdimen3\font minus \fontdimen4\font\relax}
\providecommand{\BIBforeignlanguage}[2]{{%
\expandafter\ifx\csname l@#1\endcsname\relax
\typeout{** WARNING: IEEEtran.bst: No hyphenation pattern has been}%
\typeout{** loaded for the language `#1'. Using the pattern for}%
\typeout{** the default language instead.}%
\else
\language=\csname l@#1\endcsname
\fi
#2}}
\providecommand{\BIBdecl}{\relax}
\BIBdecl

\bibitem{tedeschini2022decentralized}
B.~C. Tedeschini, S.~Savazzi, R.~Stoklasa, L.~Barbieri, I.~Stathopoulos, M.~Nicoli, and L.~Serio, ``Decentralized federated learning for healthcare networks: A case study on tumor segmentation,'' \emph{IEEE access}, vol.~10, pp. 8693--8708, 2022.

\bibitem{shiranthika2023decentralized}
C.~Shiranthika, P.~Saeedi, and I.~V. Baji{\'c}, ``Decentralized learning in healthcare: a review of emerging techniques,'' \emph{IEEE Access}, vol.~11, pp. 54\,188--54\,209, 2023.

\bibitem{palmieri2024robustness}
L.~Palmieri, C.~Boldrini, L.~Valerio, A.~Passarella, M.~Conti, and J.~Kert{\'e}sz, ``Robustness of decentralised learning to nodes and data disruption,'' \emph{arXiv preprint arXiv:2405.02377}, 2024.

\bibitem{valerio2023coordination}
L.~Valerio, C.~Boldrini, A.~Passarella, J.~Kert{\'e}sz, M.~Karsai, and G.~I{\~n}iguez, ``Coordination-free decentralised federated learning on complex networks: Overcoming heterogeneity,'' \emph{arXiv preprint arXiv:2312.04504}, 2023.

\bibitem{BKHMSILatentInterpolation}
BKHMSI, ``Mnist-latent-interpolation: Count{down, up} with mnist using latent interpolation,'' \url{https://github.com/BKHMSI/MNIST-Latent-Interpolation}, 2019, accessed: 2025-01-09.

\bibitem{mcmahan2017communication}
B.~McMahan, E.~Moore, D.~Ramage, S.~Hampson, and B.~A. y~Arcas, ``Communication-efficient learning of deep networks from decentralized data,'' in \emph{Artificial intelligence and statistics}.\hskip 1em plus 0.5em minus 0.4em\relax PMLR, 2017, pp. 1273--1282.

\bibitem{hard2018federated}
A.~Hard, K.~Rao, R.~Mathews, S.~Ramaswamy, F.~Beaufays, S.~Augenstein, H.~Eichner, C.~Kiddon, and D.~Ramage, ``Federated learning for mobile keyboard prediction,'' \emph{arXiv preprint arXiv:1811.03604}, 2018.

\bibitem{tabacof2016exploring}
P.~Tabacof and E.~Valle, ``Exploring the space of adversarial images,'' in \emph{2016 international joint conference on neural networks (IJCNN)}.\hskip 1em plus 0.5em minus 0.4em\relax IEEE, 2016, pp. 426--433.

\bibitem{bhagoji2019analyzing}
A.~N. Bhagoji, S.~Chakraborty, P.~Mittal, and S.~Calo, ``Analyzing federated learning through an adversarial lens,'' in \emph{International conference on machine learning}.\hskip 1em plus 0.5em minus 0.4em\relax PMLR, 2019, pp. 634--643.

\bibitem{sun2024gan}
W.~Sun, B.~Gao, K.~Xiong, Y.~Wang, P.~Fan, and K.~B. Letaief, ``A gan-based data poisoning attack against federated learning systems and its countermeasure,'' \emph{arXiv preprint arXiv:2405.11440}, 2024.

\bibitem{zhang2020poisongan}
J.~Zhang, B.~Chen, X.~Cheng, H.~T.~T. Binh, and S.~Yu, ``Poisongan: Generative poisoning attacks against federated learning in edge computing systems,'' \emph{IEEE Internet of Things Journal}, vol.~8, no.~5, pp. 3310--3322, 2020.

\bibitem{tolpegin2020data}
V.~Tolpegin, S.~Truex, M.~E. Gursoy, and L.~Liu, ``Data poisoning attacks against federated learning systems,'' in \emph{Computer security--ESORICs 2020: 25th European symposium on research in computer security, ESORICs 2020, guildford, UK, September 14--18, 2020, proceedings, part i 25}.\hskip 1em plus 0.5em minus 0.4em\relax Springer, 2020, pp. 480--501.

\bibitem{xia2023poisoning}
G.~Xia, J.~Chen, C.~Yu, and J.~Ma, ``Poisoning attacks in federated learning: A survey,'' \emph{IEEE Access}, vol.~11, pp. 10\,708--10\,722, 2023.

\bibitem{pham2024data}
A.~Pham, M.~Potop-Butucaru, S.~Tixeuil, and S.~Fdida, ``Data poisoning attacks in gossip learning,'' in \emph{International Conference on Advanced Information Networking and Applications}.\hskip 1em plus 0.5em minus 0.4em\relax Springer, 2024, pp. 213--224.

\bibitem{gentz2015detection}
R.~Gentz, H.-T. Wai, A.~Scaglione, and A.~Leshem, ``Detection of data injection attacks in decentralized learning,'' in \emph{2015 49th Asilomar Conference on Signals, Systems and Computers}.\hskip 1em plus 0.5em minus 0.4em\relax IEEE, 2015, pp. 350--354.

\bibitem{cao2019understanding}
D.~Cao, S.~Chang, Z.~Lin, G.~Liu, and D.~Sun, ``Understanding distributed poisoning attack in federated learning,'' in \emph{2019 IEEE 25th international conference on parallel and distributed systems (ICPADS)}.\hskip 1em plus 0.5em minus 0.4em\relax IEEE, 2019, pp. 233--239.

\bibitem{sun2022decentralized}
T.~Sun, D.~Li, and B.~Wang, ``Decentralized federated averaging,'' \emph{IEEE Transactions on Pattern Analysis and Machine Intelligence}, vol.~45, no.~4, pp. 4289--4301, 2022.

\bibitem{savazzi2020federated}
S.~Savazzi, M.~Nicoli, and V.~Rampa, ``Federated learning with cooperating devices: A consensus approach for massive iot networks,'' \emph{IEEE Internet of Things Journal}, vol.~7, no.~5, pp. 4641--4654, 2020.

\bibitem{sikar2024misclassification}
D.~Sikar, A.~Garcez, R.~Bloomfield, T.~Weyde, K.~Peeroo, N.~Singh, M.~Hutchinson, D.~Laksono, and M.~Reljan-Delaney, ``The misclassification likelihood matrix: Some classes are more likely to be misclassified than others,'' \emph{arXiv preprint arXiv:2407.07818}, 2024.

\bibitem{goodfellow2014generative}
I.~Goodfellow, J.~Pouget-Abadie, M.~Mirza, B.~Xu, D.~Warde-Farley, S.~Ozair, A.~Courville, and Y.~Bengio, ``Generative adversarial nets,'' \emph{Advances in neural information processing systems}, vol.~27, 2014.

\bibitem{schmidhuber2020generative}
J.~Schmidhuber, ``Generative adversarial networks are special cases of artificial curiosity (1990) and also closely related to predictability minimization (1991),'' \emph{Neural Networks}, vol. 127, pp. 58--66, 2020.

\bibitem{schmidhuber1990making}
------, \emph{Making the world differentiable: on using self supervised fully recurrent neural networks for dynamic reinforcement learning and planning in non-stationary environments}.\hskip 1em plus 0.5em minus 0.4em\relax Inst. f{\"u}r Informatik, 1990, vol. 126.

\bibitem{barabasi1999emergence}
A.-L. Barab{\'a}si and R.~Albert, ``Emergence of scaling in random networks,'' \emph{science}, vol. 286, no. 5439, pp. 509--512, 1999.

\bibitem{hagberg2008exploring}
A.~Hagberg, P.~J. Swart, and D.~A. Schult, ``Exploring network structure, dynamics, and function using networkx,'' Los Alamos National Laboratory (LANL), Los Alamos, NM (United States), Tech. Rep., 2008.

\bibitem{barabasi2013network}
A.-L. Barab{\'a}si, ``Network science,'' \emph{Philosophical Transactions of the Royal Society A: Mathematical, Physical and Engineering Sciences}, vol. 371, no. 1987, p. 20120375, 2013.

\bibitem{lecun1998mnist}
Y.~LeCun, ``The mnist database of handwritten digits,'' \emph{http://yann. lecun. com/exdb/mnist/}, 1998.

\bibitem{mcallister2025decentralized}
D.~McAllister, M.~Tancik, J.~Song, and A.~Kanazawa, ``Decentralized diffusion models,'' \emph{arXiv preprint arXiv:2501.05450}, 2025.

\end{thebibliography}

\end{document}